\documentclass{article}

\usepackage{PRIMEarxiv}

\usepackage[utf8]{inputenc} 
\usepackage[T1]{fontenc}    
\usepackage{hyperref}       
\usepackage{url}            
\usepackage{booktabs}       
\usepackage{amsfonts}       
\usepackage{nicefrac}       
\usepackage{microtype}      
\usepackage{lipsum}
\usepackage{graphicx}
\usepackage{amssymb}
\usepackage{algorithm}
\usepackage{algpseudocode}
\usepackage{tabularx} 
\usepackage{booktabs} 
\usepackage{cellspace} 
\usepackage{tabularray}
\usepackage{rotating}
\usepackage{diagbox}
\usepackage{subcaption}
\usepackage{multirow}
\usepackage{multicol}
\usepackage{adjustbox}
\usepackage{titlesec}
\usepackage{biblatex} 
\graphicspath{{media/}}     

\title{Hinge-FM2I: An Approach using Image Inpainting for Interpolating Missing Data in Univariate Time Series}

\author{
Saad Noufel, Nadir Maaroufi, Mehdi Najib, Mohamed Bakhouya \\
International University of Rabat \\
College of Engineering and Architecture, TICLab, LERMA Lab \\
Sala Al Jadida 11000, Morocco\\
\texttt{\{saad.noufel, nadir.maaroufi, mehdi.najib, mohamed.bakhouya\}@uir.ac.ma} \\
}

\begin{document}
\maketitle

\begin{abstract}
	Accurate time series forecasts are crucial for various applications, such as traffic management, electricity consumption, and healthcare. However, limitations in models and data quality can significantly impact forecasts accuracy. One common issue with data quality is the absence of data points, referred to as missing data. It is often caused by sensor malfunctions, equipment failures, or human errors. This paper proposes Hinge-FM2I, a novel method for handling missing data values in univariate time series data. Hinge-FM2I builds upon the strengths of the Forecasting Method by Image Inpainting (FM2I). FM2I has proven effective, but selecting the most accurate forecasts remain a challenge. To overcome this issue, we proposed a selection algorithm. Inspired by door hinges, Hinge-FM2I drops a data point either before or after the gap (left/right-hinge), then use FM2I for imputation, and then select the imputed gap based on the lowest error of the dropped data point. Hinge-FM2I was evaluated on a comprehensive sample composed of 1356 time series, extracted from the M3 competition benchmark dataset, with missing value rates ranging from 3.57\% to 28.57\%. Experimental results demonstrate that Hinge-FM2I significantly outperforms established methods such as, linear/spline interpolation, K-Nearest Neighbors (K-NN), and ARIMA. Notably, Hinge-FM2I achieves an average Symmetric Mean Absolute Percentage Error (sMAPE) score of 5.6\% for small gaps, and up to 10\% for larger ones. These findings highlight the effectiveness of Hinge-FM2I as a promising new method for addressing missing values in univariate time series data.
\end{abstract}

\keywords{Univariate Time Series \and Missing data imputation \and Interpolation \and Image Inpainting \and Artificial Intelligence}

\section{Introduction}
In today's data-driven world, forecasting future values of a time series when real-time data points come progressively through time, called online time series forecasting \cite{pham2022learning}, has become a necessary tool for extracting useful insights from the continuous flow of information. In this sense, the model is continuously updated as new data observations become available. This allows the model to adapt to changes in the underlying time series and to make more accurate forecasts. This forecasting approach is becoming increasingly important across a variety of domains, such as traffic management \cite{malek2023helecar}, electricity consumption \cite{hadri2021performance}, healthcare \cite{hyndman2018forecasting}, weather data \cite{fiore2018road}, transportation \cite{malek2021multivariate}, and energy management \cite{elmouatamid2020mapcast}. However, the use of real-time data presents its own set of challenges, such as data availability \cite{desai2019real}, data quality \cite{kirkendall2019data}. Some of the issues of real-time data quality include the potential to introduce noises, anomalies, or missing data \cite{guo2016robust}. This latter, also referred to as missing values, occurs when there is no data stored for certain variables \cite{emmanuel2021survey}. In other terms, missing data refers to the absence of observations that are expected to be present. Missing data can happen for various reasons. For instance, in the context of Internet of Things (IoT), missing data can occurs due to various reasons, such as sensor malfunction, network failure, or data transmission errors \cite{liu2020missing}. While, in the context of particulate air pollutant, it can occurs due to power supply failures, problems with air aspiration pumps, or electronic processing malfunction \cite{shakya2023selection}.

A large gap of missing data, of size $T$, is an interval $[t:t + T -1[$ containing consecutive missing values. It poses significant challenges for online forecasting models as they introduce substantial information loss, making it difficult to accurately predict future values \cite{wu2022data}. To address this challenge, two main approaches are commonly employed to handle large gaps of missing data: \textit{deletion} or \textit{imputation}. The first one involves simply ignoring observations with missing values. Deletion approach consists of two techniques, listwise deletion, which discards all observations with missing values for any variable of interest, and pairwise deletion, which does not exclude the entire unit but uses as much data as possible. However, a major drawback of the deletion approach is losing that it can significantly reduce the size of the dataset, which leads to a loss of information \cite{wafaa2022comparison}. The imputation approach estimates the missing values based on available information and underlying patterns. Multiple techniques, such as mean, median, last observation carried forward (LOCF), and interpolation techniques, are frequently used to fill in missing data in univariate datasets \cite{peugh2004missing}, \cite{pratama2016review}, \cite{osman2018survey}. While these methods can handle a few consecutive missing values or short data gaps, they are not well-suited for dealing with multiple types of data gaps, particularly large ones \cite{zainuddin2022time}. While common approaches like deletion and imputation have limitations in handling large data gaps, some researchers have proposed leveraging data from nearby sensors or devices experiencing similar conditions. For instance, \cite{guastella2021edge} proposed a method combining Multi-Agent Systems and IoT to impute missing values by using mobile intermittent IoT devices and local computations in areas where sensors perceive analogous dynamics. This spatial approach aims to better estimate missing information in large-scale data analysis with complex, multiple data gaps.

In this research paper, we propose a novel imputation method to address large gaps of missing data. The technique begins by identifying the gap and dropping/storing one value before or after the gap. The resulting time series is then fed into a forecasting method based on image inpainting (FM2I) \cite{maaroufi2021predicting}. FM2I transforms the time series into an image, applies a mask, and employs a patch-based image inpainting approach to fill the masked region. Once, the image is inpainted it is converted again to a matrix, a selection algorithm is then applied to select the most accurate forecasts. Unlike the original FM2I, which focuses on forecasting (extrapolation), our objective is to impute missing data through interpolation. To adapt FM2I for this purpose, we propose a simple shift in the mask position, allowing us to leverage its image inpainting capabilities for interpolation. We introduce a selection algorithm, named Hinge-FM2I, which compares the dropped value with the generated value and then selects the sequence with the minimal difference error. To showcase the efficiency of our approach, an extensive evaluation was conducted on a diverse sample extracted from the M3 benchmark dataset. The obtained results demonstrate that Hinge-FM2I significantly outperforms existing techniques in imputing large gaps of missing data.

In summary, the main general contributions of this work are two folds:
\begin{itemize}
	\item Introduce Hinge-FM2I to impute large gaps of missing data based on the forecasting by image inpainting.
	\item Extensive experiments are conducted to evaluate the proposed method and compare it against different commonly used imputation methods as well as real recorded observations (ground truth).
\end{itemize}

The reminder of this paper is structured as follow. Section 2 provides a comprehensive state-of-the-art review of the methods handling missing data in univariate time series. Section 3 introduces Hinge-FM2I for imputing large gaps of missing data in univariate time series. Section 4 reveals our results showcasing the effectiveness of our approach compared to existing techniques. Section 5 summarizes our findings, while highlighting perspectives.

\section{Related work}
Real-time forecasting refers to the process of predicting future values of a time series as new data becomes available. This approach involves continuously updating the model with the latest observations and using it to make predictions for the next time horizon \cite{gajamannage2023real}. It holds utmost importance in today's data-driven world, but is faced with data quality problems. Among these problems, the existence of missing data. It causes the reliability and accuracy of forecasting models to heavily drop. Large gaps of missing data can be identified as sequences of consecutive data points that are missing. The size of the gap is determined by the number of consecutive missing data points. Beginning at a specific point, $x_t$, and extending to $x_{t+T-1}$, where $T$ represents the size of this gap. The missing data in these gaps can be classified into three types, missing completely at random (MCAR), missing at random (MAR) and Not missing at random (NMAR) \cite{rubin1976inference}. It is essential to know why data is missing in order to choose a good method to deal with those gaps. Understanding the reasons behind the missing data can help pick the right imputation technique \cite{moritz2015comparison}. But in reality, figuring out those reasons is frequently difficult, particularly when there's no information about the missing data at all, or when the missing data follows a complicated pattern \cite{gomez2014practical}. To the best of our knowledge, most current research works focus on the three defined missing data types. The following sections presents the employed methods to deal with missing data in univariate time series.

Missing data in univariate time series can be dealt with following two main approaches: \textit{deletion} or \textit{imputation}. On one hand, the deletion approach entirely removes observations with missing values from the datasets. While this approach is straightforward and easy to implement, it often leads to substantial information loss, especially when large gaps of missing data are present \cite{wafaa2022comparison}. On the other hand, imputation techniques are employed to estimate the missing values based on the available information and the underlying patterns in the time series. Imputation offers the advantage of retaining the full datasets, thereby preserving valuable information while maintaining the continuity of the time series \cite{wafaa2022comparison}. However, the extra uncertainty added by the imputation process itself is usually not taken into account. This means the calculated variance of the estimated relationships in the data is typically underestimated and appears smaller than it really is. Another key issue with imputation methods, especially the simpler types, is that they can produce biased or inaccurate estimates of the associations or connections between variables in the data \cite{junger2015imputation}.
\begin{figure}[h]
	\centering
	\includegraphics[width=\textwidth]{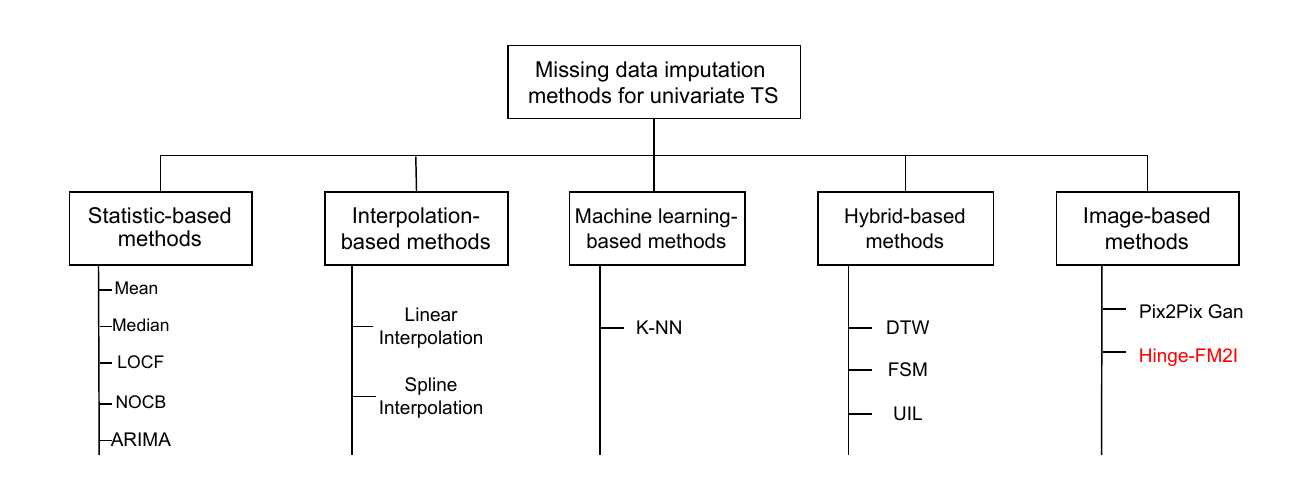}
	\caption{Comprehensive Overview of missing data imputation methods.}
	\label{fig:methods}
\end{figure}

In the past decade, several methods have been proposed to deal wiht missing data in univariate time series. These methods can be classified into four sub-categories, as depicted in Fig.\ref{fig:methods}: \textit{Statistical-based}, \textit{Interpolation-based}, \textit{Machine learning-based}, and \textit{Image-based}. Firstly, statistical-based methods estimate and replace missing values with statistical measures, such as mean, median, LOCF, next observation carried backward (NOCB), or ARIMA \cite{little2019statistical}. While these methods are simple and easy to implement with a fast computation time. However, these methods may not work well when a large amount of data is missing \cite{phan2020machine}. For instance, these methods underestimates the variance, ignores the correlation between the features, thus often leads to poor imputation \cite{bertsimas2018predictive}. Secondly, interpolation-based methods such as, linear and spline interpolation, creates a continuous function that approximates the relationship between the observed data points. On one hand, linear interpolation constructs a straight line between two known data points and estimates the missing value based on this line. On the other hand, spline interpolation uses piecewise polynomial functions to create a smooth curve that passes through the known data points, providing a more accurate estimation of missing values \cite{noor2014filling}. Even though it is easier to implement, linear interpolation assumes a linear relationship between data points, which might not always reflect the true underlying patterns in the time series data, meanwhile, spline interpolation does not consider the underlying structure of the data, which can result in inaccurate imputations \cite{zhang2016missing}. Thirdly, machine learning-based methods deals with missing data by creating a predictive model to estimate values that will substitute the missing items. By learning patterns and relationships from the available data, it allows them to make predictions and fill in missing values based on the existing information in the dataset. For instance, k-Nearest Neighbor (kNN) \cite{dubey2021efficient} is widely used to impute missing data in univariate time series.

While traditional imputation techniques like deletion and interpolation methods have been widely used, they often fall short when dealing with complex patterns or large gaps in univariate time series data. To address these limitations, researchers have proposed more advanced state-of-the-art hybrid- approaches. One class of methods focuses on aligning and matching the observed and missing portions of the time series. For instance, the authors in \cite{caillault2020dynamic} proposed, Dynamic Time Warping (DTW), an imputation technique that compares two time series by aligning them in a way that minimizes the differences between corresponding points. DTW-based imputation works by finding the best alignment between the observed and missing parts of the time series, and then using this alignment to estimate the missing value. However, the quality of the imputed values depends on the accuracy of the alignment. Similarly, \cite{khampuengson2023novel} introduced Full Subsequence Matching (FSM), which identifies the most similar sub-sequence from historical data and adapts it to fit the missing part. Nevertheless, FSM may not perform well if there are too many missing values or if historical patterns are not representative. Another approach involves combining univariate and multivariate imputation techniques, as proposed in \cite{velasco2022novel} for marine machinery systems. While this framework shows promise in that domain, potentially its applicability and limitations in other contexts are not extensively discussed, limiting the generalization of the findings. In their work, \cite{han2023univariate} proposed a univariate imputation method (UIM) specialized for wastewater treatment processes. The UIM approach consists of three main steps. Firstly, missing gaps in the data are initially filled using linear interpolation. Secondly, the UIM applies a seasonal trend decomposition procedure based on locally estimated scatterplot smoothing (loess), also known as STL. After decomposing the time series, the trend and remainder components are imputed separately using support vector regression (SVR), while the seasonal component is imputed through self-similarity decomposition (SSD). In the final step, the imputed components are recombined by summing them, yielding the imputed time series. 

\noindent Recently forecasting by image inpainting (FM2I) \cite{maaroufi2021predicting} was proposed to extrapolate time series. But when it comes to interpolation of missing data in univariate time series, to the best of our knowledge, the only approach that tackles this issue was proposed in \cite{almeida2023univariate}. They proposed using a conditional generative adversarial network (Pix2Pix GAN) to impute missing data by transforming time series into images. However, the performance of this method may depend on the selected network parameters. To overcome this limitation, Hinge-FM2I is proposed for imputing univariate time series based on image inpainting.

\section{Hinge-FM2I approach}
\begin{figure}[htbp]
	\centering
	\begin{subfigure}{0.35\textwidth}
		\centering
		\includegraphics[width=\textwidth]{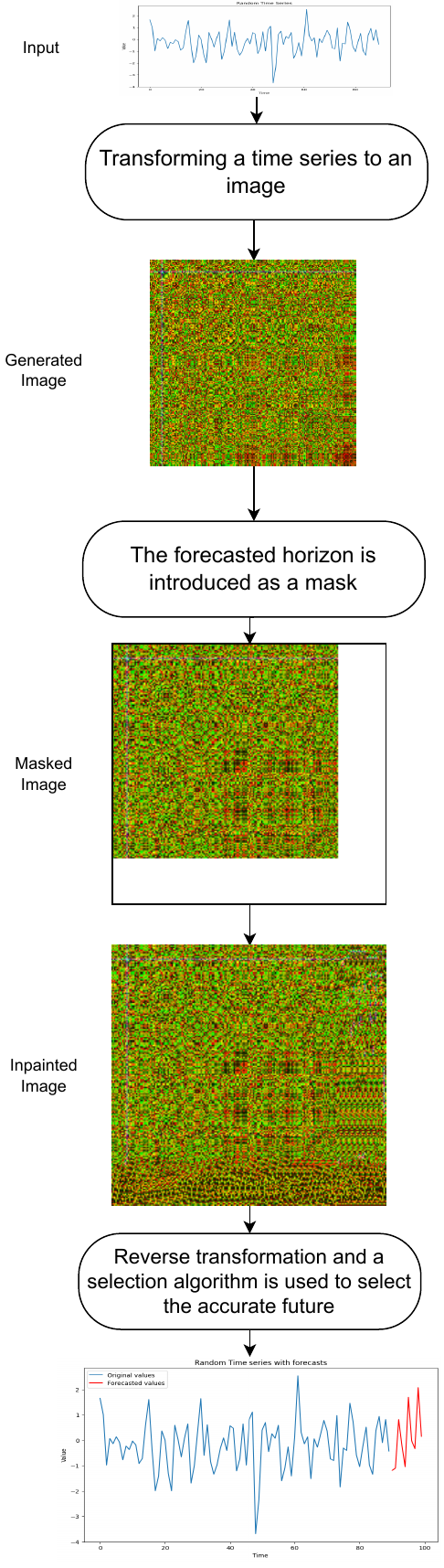}
		\caption{}
		\label{fig:figure1}
	\end{subfigure}
	\hfill
	\begin{subfigure}{0.45\textwidth}
		\centering
		\includegraphics[width=\textwidth]{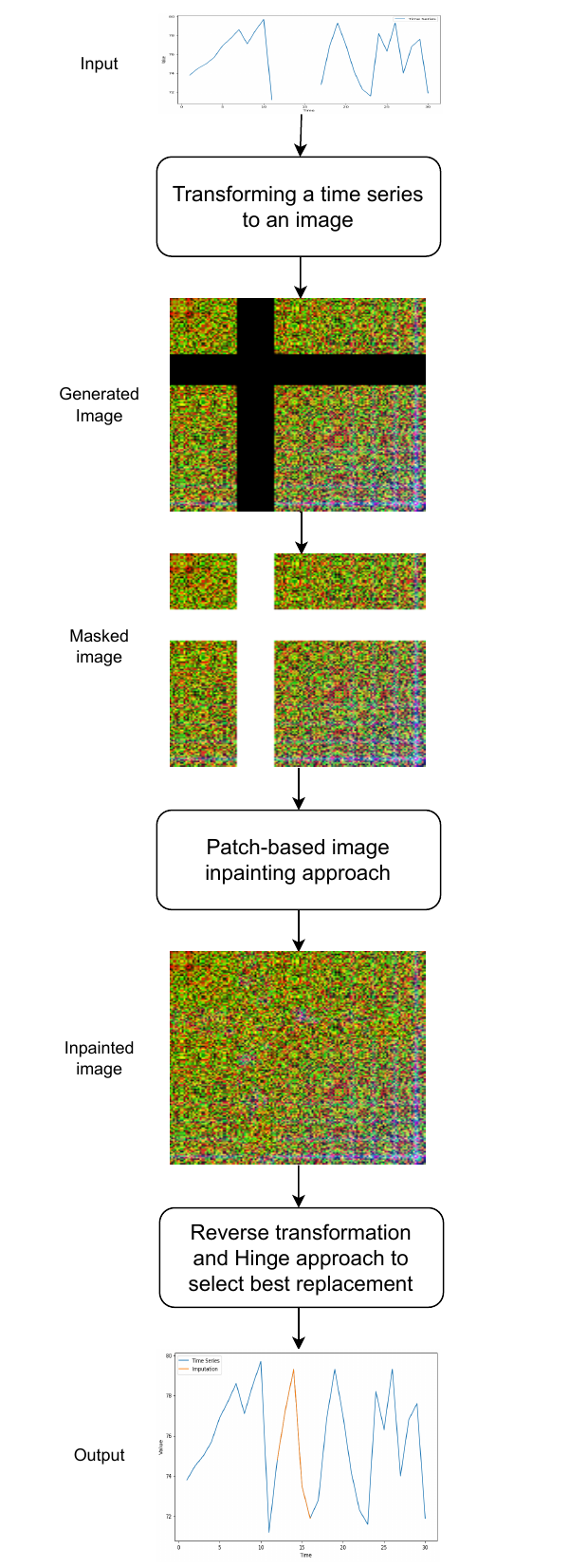}
		\caption{}
		\label{fig:figure2}
	\end{subfigure}
	\caption{FM2I and Hinge-FM2I framework}
	\label{fig:two_figures}
\end{figure}

Dealing with missing data in univariate time series is a significant challenge since missing data make it difficult to accurately forecast future values. The above mentioned methods may fail to capture complex underlying patterns, leading to inaccurate estimates or loss of valuable information, especially when dealing with large gaps of missing data. To address these limitations, we propose Hinge-FM2I, a new method for filling in missing values in univariate time series data. Hinge-FM2I is an extension of FM2I \cite{maaroufi2021predicting}. It is worth noting that, FM2I algorithm can be structured into three steps as depicted in Fig.\ref{fig:figure1}. The first step consists of all the steps required for preparing the time series. It consists of rescaling to either $[-1,1]$ or $[0,1]$. Once a rescaled time series is calculated, noted $res\_TS$, it is transformed to its corresponding image. The second step consists of extrapolating and inpainting the generated images using a patch-based algorithm. Note that the mask correspond to the forecasting horizon. In the third and last step, the reverse process is applied in order to both extract the forecast values from the image and reverse all the scaling steps.

Although experiments showed great results in terms of forecasting accuracy, FM2I has some limitations. First, the patch-based image inpainting approach uses existing data to choose the the most suitable pixel to inpaint, but with large gaps of missing values in the time series data, the generated image is affected which might lead the patch-based approach to introduce mismatch error. Second, choosing the accurate forecast for extrapolation or best imputation for interpolation requires defining a new selection algorithm. Hinge-FM2I introduces a selection technique based on the idea of door hinges, aiming to select the best sequence from a matrix of possible imputation sequences. 

\begin{algorithm}
	\caption{Hinge-FM2I Algorithm for Time Series Imputation}
	\label{fig:hinge-fm2i-algorithm}
	\begin{algorithmic}
		\Require Time series with a sequence of missing values, denoted as $TS$
		\Ensure Imputed time series with no missing values
		\State $\setminus \setminus$ \textbf{Step 1:} Introduce missing data
		\State $[New\_Ts,\, Gap, Xp] \gets$ Replace\_Real\_Values($TS$, $idx$, $size$)
		\For {matrix $\in \{ 1,2,3,4,5,6 \}$}
		\State $\setminus \setminus$ \textbf{Step 2:} Time series rescalling 
		\State $Ts\_scaled \gets$ minMaxScaling($New\_TS$, matrix) 
		\State $\setminus \setminus$ \textbf{Step 3:} Matrix generation
		\State $Gene\_Mat \gets$ gen\_Mat($Ts\_scaled$, matrix)
		\State $\setminus \setminus$ \textbf{Step 4:} Image generation
		\State $Gene\_Mat\_256 \gets$ gen\_Mat\_256($Gene\_Mat$)
		\State $Gene\_Img \gets$ gen\_Img($Gene\_Mat\_256$)
		\State $\setminus \setminus$ \textbf{Step 5:} Inpainting process
		\State $Inpainted\_Img \gets$ in\_paint($Gene\_Img$)
		\State $\setminus \setminus$ \textbf{Step 6:} The reverse transformation
		\State $Imputed\_Gaps \gets$ Ts\_from\_Image($Inpainted\_Img$, matrix)
		\State $\setminus \setminus$ \textbf{Step 7:} Hinge Selection
		\State $Best\_Gap \gets$ Best\_Choice($Imputed\_Gaps$)
		\State [$sMape$, $RMSE$, $MAE$] $\gets$ accuracyEval($Best\_Gap$, $Gap$)
		\State $Ts\_Imputed \gets$ Replace\_Nan($Ts\_scaled$, $Best\_Gap$)
		\State $Imputed\_Ts \gets$ reverseScaling($Ts\_Imputed$)
		\EndFor
	\end{algorithmic}
\end{algorithm}

The Hinge-FM2I algorithm consists of seven main steps as depicted in the Fig.\ref{fig:figure2}, and detailed in Algorithm \ref{fig:hinge-fm2i-algorithm}. The \textit{first step} deliberately introduces missing data by sequentially removing and saving data points at specific gap lengths, such as 5, 10, and 20. This allows us to simulate different types of missing data, commonly found in real-world time series datasets. In the \textit{second step}, the time series data is rescaled to the range of 0 to 1 using the min-max normalization. This involves subtracting the minimum value from each data point and then dividing by the difference between the minimum and maximum values, ensuring that all of the data points fall within the 0 to 1 range. The \textit{third step} follows the transformation described in \cite{maaroufi2021predicting}, which converts the one-dimensional time series into a two-dimensional image representation. This transformation uses six different matrices, resulting in a visual representation that captures both temporal and correlation information. The \textit{fourth step} involves using a bijective function to encode the scaled matrix values into a trio of distinct RGB values, each ranging from 0 to 255 ($256^3$ possibilities in total). After successfully encoding the data, the calculated RGB values are used to create an image that represents the original matrix, completing the transformation process. In the \textit{fifth step}, the missing values are masked, and an adapted patch-based algorithm \cite{criminisi2004region} is applied to interpolate and inpaint the generated images. The \textit{sixth step} involves retrieving all possible imputed values from the inpainted image, resulting in a single matrix containing potential imputation sequences.

The \textit{seventh step} consists of applying the hinge algorithm to the resulting matrix. This idea was inspired by replacing a door with hinges. First, the algorithm stores and replaces with NaN value the first data point either before or after the gap of missing data within the time series. For example, if the gap of size $N$ starts at the data point denoted $x_p$ until $x_{p+N-1}$, the hinge algorithm will store the value of $x_{p-1}$ and change its value to NaN. Once the inpainting process is complete, it calculates the difference between $x_{p-1}$ and its estimate $\hat{x}_{p-1}$, and selects the gap $\hat{x}_p$ till $\hat{x}_{p+N-1}$ with the minimum difference error. The time series is then imputed and reverse scaled. With this step, the proposed Hinge-FM2I method replaces large gaps of missing data.

\section{Experimental results}
\subsection{Datasets, methods and metrics}
In order to assess the effectiveness of Hinge-FM2I, we have taken a sample from the M3 competition \cite{makridakis2000m3}, we only took into consideration time series with a minimum of 70 observations. The sample dataset includes 1356 time series, which are categorized according to their types (e.g., industry, finance) and time intervals (e.g., Monthly, quarterly, yearly). We also used two public waste water treatement process public datasets to compare against state of the art approach, the first dataset can be found in (UCI) and it is a collection of waste water treatment plant in Spain, while the other is simply $NH_4$ concentration and can be obtained from imputeTS package in R software.

\noindent For the sake of evaluating the performance of the proposed method. Firstly, we compared with nine other existing methods for univariate time series (mean, median, LOCF, NOCB, LI, SI, KNN, ARIMA, DTWBI). All these methods are implemented using Python language, except the DTWBI which exists as a package in R language. The spline interpolation method employed a cubic polynomial function, whereas for the KNN approach, a grid search was performed to determine the optimal number of clusters. Once the imputation of missing values is completed, we evaluate the performance of our method, and compare it with the above stated approaches based on three different metrics. First, Symmetric Mean Absolute Percentage Error (sMAPE) measures the percentage difference between predicted and actual values, taking into account the magnitude of the values being compared. It is expressed in percentage. Second, Root Mean Squared Error (RMSE) measures the average squared difference between predicted values and actual values, with the square root taken to obtain the final measure. Third, Mean Absolute Error (MAE) measures the average absolute difference between predicted values and actual values. It provides a straightforward measure of the average prediction error. Secondly, we compared our approach to a SOTA method \cite{han2023univariate}, and evaluated the results using two indicators RMSE and similarity (Sim).
\subsection{Results and discussion}
Experiments have been conducted on a computer composed of the following characteristics, AMD Ryzen 7 3750H, 2.30 GHz, 12.00 GB RAM, x64 based processor, Windows 10. For the first experience, we have considered three missing data sequence lengths, 5, 10, and 20. These gaps sizes simulated a missing data ranging from 3.57\% to 28.57\%. On the other hand, for the second experience, we considered four missing ratio 5\%, 10\%, 15\%, and 20\%. The result and discussion section is divided into two subsection, first a numerical analyze of the imputation is presented and discussed, then a visual analysis is showcased.

\begin{figure}[htbp]
	\centering
	\begin{subfigure}[b]{0.49\textwidth}
		\includegraphics[width=\textwidth]{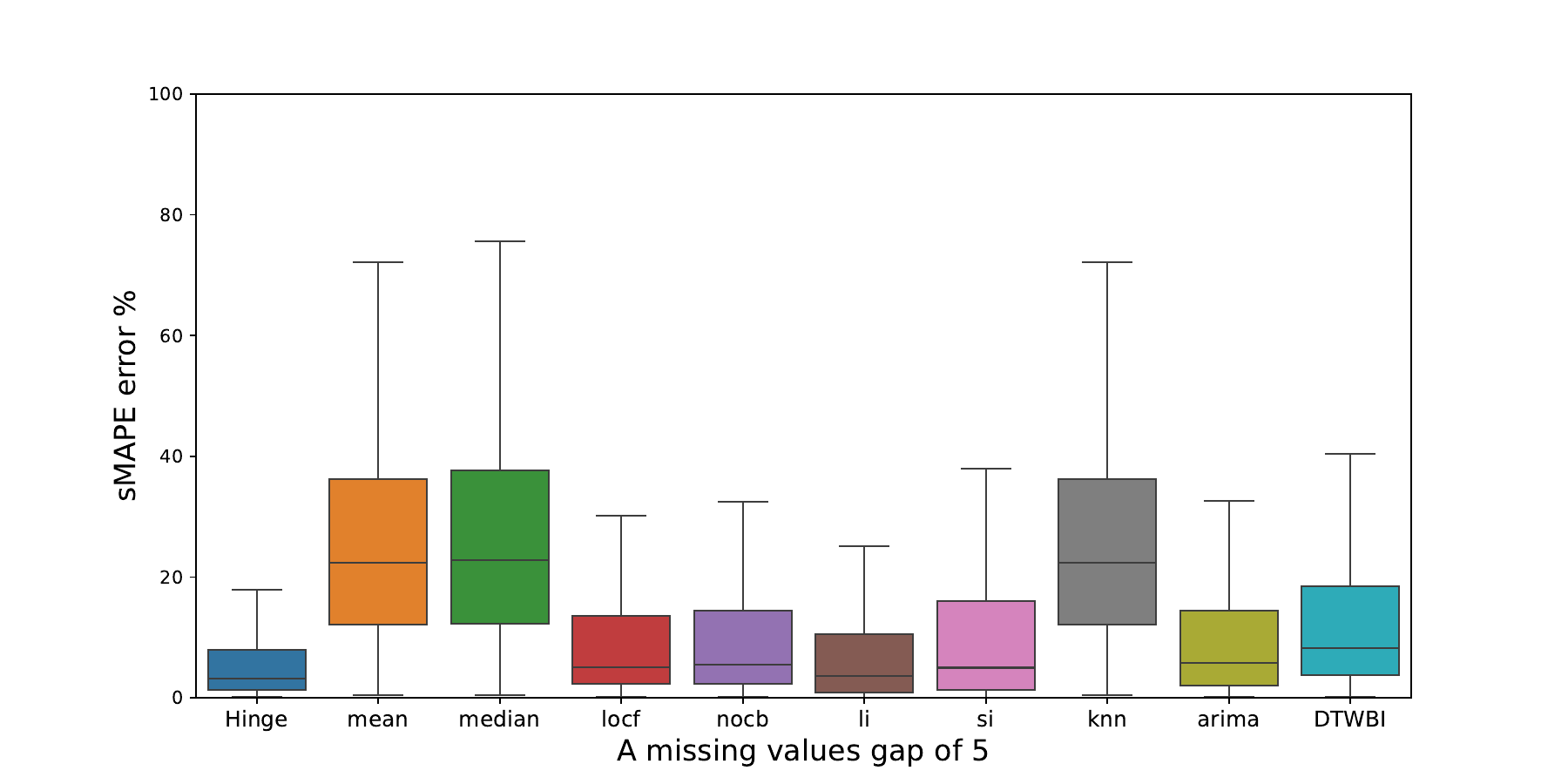}
		\caption{}
		\label{fig:BoxPlot1}
	\end{subfigure}
	\hfill
	\begin{subfigure}[b]{0.49\textwidth}
		\includegraphics[width=\textwidth]{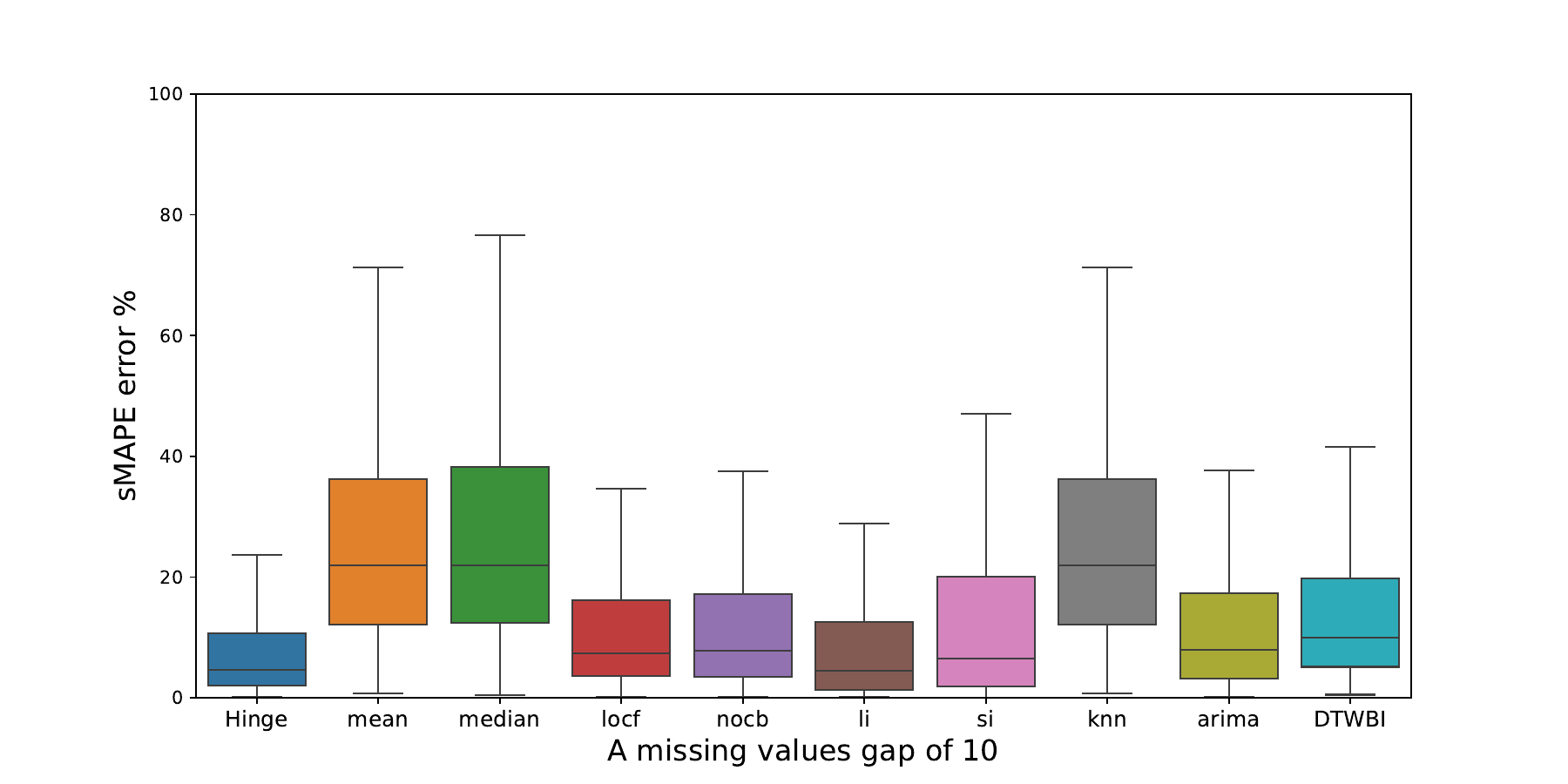}
		\caption{}
		\label{fig:BoxPlot2}
	\end{subfigure}
	\vfill
	\begin{subfigure}[b]{0.5\textwidth}
		\includegraphics[width=\textwidth]{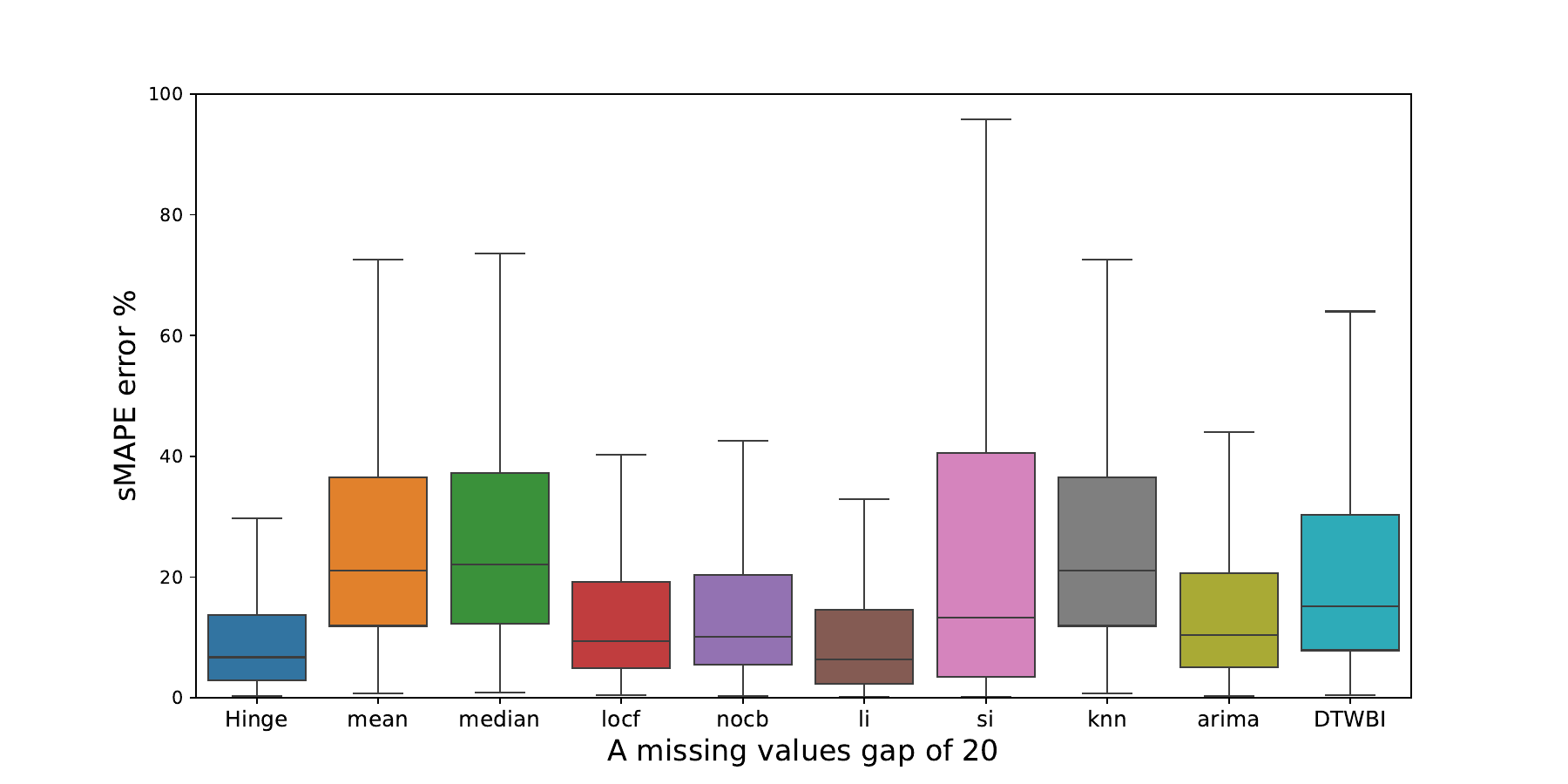}
		\caption{}
		\label{fig:BoxPlot3}
	\end{subfigure}
	\caption{Box-plots of sMAPE for various imputation methods with different size of missing gaps.}
	\label{fig:BoxPlots}
\end{figure}

\noindent In Table \ref{tab:1} the comparison, in terms of imputation accuracy, of the Hinge-FM2I against the commonly used methods. Showcase that Hinge-FM2I outperforms all the other methods in terms of all metrics. Apart from our method, only the LI method had good results in terms of sMAPE error. To better understand the effectiveness of the Hinge-FM2I, an analysis of Fig.\ref{fig:BoxPlot1} demonstrate the Hinge-FM2I approach compared to other methods. Its box plot shows a relatively low median, indicating consistently lower sMAPE values, which reflects higher prediction accuracy. In contrast, methods like "Mean" and "Knn" display larger interquartile ranges and higher median values, suggesting greater variability and lower accuracy. The "Median", "Locf", and "Nocb" approaches present moderate performance with a somewhat symmetrical distribution, but they do not match the accuracy and consistency of the Hinge-FM2I method. "Li" and "Si" methods exhibit significant variability and right skewness, indicating less reliable performance. The "Arima" method, while having a slightly left-skewed distribution, still shows a higher median than "Hinge-FM2I". In Fig.\ref{fig:BoxPlot2}, the Hinge-FM2I approach again demonstrates superior performance based on the sMAPE metric. The box plot reveals that the Hinge-FM2I column has a centrally located median with a narrow interquartile range, indicating high accuracy and consistency. Other approaches show varied performance, with some methods exhibiting higher median sMAPE values and broader interquartile ranges, reflecting lower accuracy and greater variability. The overall distribution patterns of these methods highlight their inferiority compared to the Hinge-FM2I approach. The absence of significant outliers further underscores the robustness of the Hinge-FM2I method, making it the preferred choice for achieving lower sMAPE values and, consequently, better predictive performance. The evaluation of Fig.\ref{fig:BoxPlot3} through sMAPE metric results further solidifies the superiority of the Hinge-FM2I approach. The box plot for Hinge-FM2I displays a central median within a narrow interquartile range, signifying low sMAPE values and high prediction accuracy. In comparison, other methods show wider interquartile ranges and higher medians, indicative of lower accuracy and increased variability. Some methods, despite having a symmetrical distribution, still do not achieve the same level of performance as Hinge-FM2I. The presence of more significant variability and skewness in the alternative methods emphasizes the consistency and reliability of the Hinge-FM2I approach., proposed approach, time series with different features, such as trend or seasonality with various sizes, where taken as a sample and imputed. The results of this experiment were presented in Table \ref{Tab:Diff}. As shown in this Table, the Hinge-FM2I method, in its current version, showcased great results, even across different types of time series. In summary, the Hinge-FM2I outperforms all the most used methods in terms of accuracy. It is also worth nothing that, Hinge-FM2I doesn't take into consideration neither the type of missing data nor the category of the time series.

\noindent We then evaluated the performance of our proposed method against the current state-of-the-art approach from \cite{han2023univariate}. As shown in Table \ref{Tab:UIM}, our method generally outperformed the previous state-of-the-art. For the PH.E time series, our approach achieved a lower RMSE while maintaining equal similarity (SIM) values compared to the baseline. On the SED.P dataset, our method demonstrated better similarity results with equal RMSE performance. For the SS.S and SED.S time series, our approach yielded higher similarity scores, though there was a trade-off in terms of increased RMSE values. With the RD.DBO.P data, we obtained lower RMSE but equivalent similarity to the previous work. Finally, on the $NH_4$ time series, our method improved RMSE performance at the cost of reduced similarity compared to the state-of-the-art. Overall, the results summarized in Table \ref{Tab:UIM} highlight the competitive performance of our proposed approach across a range of time series benchmarks. While trade-offs between accuracy and similarity were observed for certain datasets, our method consistently matched or exceeded state-of-the-art performance in one of the two key evaluation metrics.
\begin{table}[H]
	\centering
	\begin{tblr}{
			row{odd} = {c},
			row{4} = {c},
			row{6} = {c},
			row{8} = {c},
			row{10} = {c},
			row{12} = {c},
			cell{1}{1} = {r=2}{},
			cell{1}{2} = {c=3}{},
			cell{1}{5} = {c=3}{},
			cell{1}{8} = {c=3}{},
			hline{1,3,14} = {-}{},
			hline{2} = {2-10}{},
		}
		Methods     & Gap size of 5 &                 &                 & Gap size of 10 &                 &                 & Gap size of 20 &                 &                 \\
		& sMAPE         & RMSE            & MAE             & sMAPE          & RMSE            & MAE             & sMAPE          & RMSE            & MAE             \\
		Left-Hinge  & 5.94          & 298.28          & 235.70          & \textbf{8.23}  & \textbf{461.18} & \textbf{349.31} & \textbf{10.32} & \textbf{602.87} & \textbf{458.23} \\
		Right-Hinge & \textbf{5.69} & \textbf{285.56} & \textbf{223.69} & 8.98           & 492.87          & 368.95          & 12.32          & 643.31          & 488.33          \\
		Mean        & 27.69         & 1392.29         & 1314.47         & 27.41          & 1374.47         & 1281.95         & 27.31          & 1399.82         & 1284.98         \\
		Median      & 28.65         & 1430.75         & 1350.45         & 28.48          & 1419.84         & 1324.00         & 28.47          & 1448.61         & 1331.56         \\
		LOCF        & 10.68         & 587.92          & 499.60          & 12.93          & 687.41          & 570.39          & 14.93          & 826.50          & 685.75          \\
		NOCB        & 11.99         & 629.39          & 531.70          & 13.76          & 706.33          & 587.85          & 16.16          & 846.82          & 699.50          \\
		LI          & 8.23          & 472.04          & 385.84          & 10.13          & 544.04          & 437.10          & 10.88          & 613.97          & 488.65          \\
		SI          & 14.97         & 677.85          & 586.02          & 19.69          & 915.10          & 785.89          & 31.23          & 1637.59         & 1398.90         \\
		KNN         & 27.69         & 1392.29         & 1314.47         & 27.41          & 1374.47         & 1281.95         & 27.31          & 1399.82         & 1284.98         \\
		ARIMA       & 16.90         & 803.17          & 711.87          & 20.48          & 897.30          & 776.31          & 25.57          & 1054.97         & 905.99          \\
		DTWBI       & 14.40         & 745.5           & 630.30          & 15.25          & 815.60          & 710.20          & 21.97          & 980.40          & 835.70          
	\end{tblr}
	\caption{Comparison of Imputation Accuracy Metrics}
	\label{tab:1}
\end{table}

\begin{table}
	\centering
	\begin{adjustbox}{width=\textwidth,center}
		\begin{tblr}{
				cell{1}{1} = {r=2}{},
				cell{1}{2} = {r=2}{t},
				cell{1}{3} = {c=2}{},
				cell{1}{6} = {c=2}{},
				cell{1}{9} = {c=2}{},
				cell{1}{12} = {c=2}{},
				cell{1}{15} = {c=2}{},
				cell{1}{18} = {c=2}{},
				cell{3}{1} = {c},
				cell{4}{1} = {c},
				cell{5}{1} = {c},
				cell{6}{1} = {c},
				hline{1,3,7} = {-}{},
				hline{2} = {3-4,6-7,9-10,12-13,15-16,18-19}{},
			}
			{Missing~\\ratio } & Method                         & PH.E                        &                    &  & SED.P               &                    &  & SS.S                &                    &  & SED.S               &                    &  & RD.DBO.P           &                    &  & NH4                 &                    \\
			&                                & RMSE                        & SIM                &  & RMSE                & SIM                &  & RMSE                & SIM                &  & RMSE                & SIM                &  & RMSE               & SIM                &  & RMSE                & SIM                \\
			5\%                & {Left-Hinge\\Right-Hinge\\UIM} & {\textbf{0.09}\\0.11\\0.13} & {\textbf{0.86}\\0.82\\0.82} &  & {\textbf{0.05}\\0.08\\\textbf{0.05}}  & {\textbf{0.85}\\0.80\\0.79} &  & {0.83\\0.71\\\textbf{0.05}}  & {0.86\\\textbf{0.90}\\0.76} &  & {0.36\\0.36\\\textbf{0.012}} & {\textbf{0.91}\\\textbf{0.91}\\0.75} &  & {\textbf{0.13}\\\textbf{0.13}\\0.16} & {0.81\\\textbf{0.82}\\0.80} &  & {\textbf{0.04}\\0.12\\0.08}  & {0.73\\0.60\\\textbf{0.91}} \\
			10\%               & {Left-Hinge\\Right-Hinge\\UIM} & {\textbf{0.07}\\0.16\\0.14} & {\textbf{0.85}\\0.80\\\textbf{0.85}}   &  & {\textbf{0.05}\\0.06\\\textbf{0.05}} & {\textbf{0.84}\\0.81\\0.82} &  & {0.64\\0.56\\\textbf{0.08}}  & {0.90\\\textbf{0.91}\\0.82}          &  & {0.25\\0.26\\\textbf{0.010}} & {\textbf{0.95}\\\textbf{0.95}\\0.82} &  & {0.20\\\textbf{0.13}\\0.19}          & {0.80\\\textbf{0.83}\\\textbf{0.83}} &  & {\textbf{0.03}\\0.11\\0.102} & {0.78\\0.60\\\textbf{0.89}} \\
			15\%               & {Left-Hinge\\Right-Hinge\\UIM} & {\textbf{0.13}\\0.15\\0.16} & {\textbf{0.85}\\0.83\\\textbf{0.85}} &  & {0.06\\0.08\\\textbf{0.048}}         & {\textbf{0.85}\\0.81\\0.84} &  & {0.50\\0.52\\\textbf{0.084}} & {\textbf{0.91}\\\textbf{0.91}\\0.84} &  & {0.21\\0.21\\\textbf{0.011}} & {0.96\\\textbf{0.97}\\0.77}          &  & {0.23\\\textbf{0.17}\\0.19}          & {0.81\\\textbf{0.84}\\0.83}          &  & {\textbf{0.07}\\0.14\\0.102} & {0.71\\0.69\\\textbf{0.90}} \\
			20\%               & {Left-Hinge\\Right-Hinge\\UIM} & {\textbf{0.08}\\0.14\\0.17} & {0.84\\0.83\\\textbf{0.86}}          &  & {\textbf{0.05}\\0.08\\\textbf{0.05}} & {\textbf{0.85}\\0.81\\0.84} &  & {0.46\\0.47\\\textbf{0.082}} & {0.89\\\textbf{0.91}\\0.84}          &  & {0.18\\0.18\\\textbf{0.011}} & {\textbf{0.97}\\\textbf{0.97}\\0.82} &  & {\textbf{0.21}\\0.24\\\textbf{0.21}} & {\textbf{0.82}\\0.79\\\textbf{0.82}} &  & {\textbf{0.10}\\0.16\\0.13}  & {0.68\\0.63\\\textbf{0.85}} 
		\end{tblr}
	\end{adjustbox}	
	\caption{Comparison of our approach to SOTA on six time series}
	\label{Tab:UIM}
\end{table}

In Table \ref{Tab:Diff}, we present the evaluation results that demonstrate the robustness of our proposed approach, Hinge-FM2I, when applied to time series data with varying characteristics. Specifically, we consider two types of time series: stationary and non-stationary. To determine the stationarity of the time series, we employed the Augmented Dickey-Fuller (ADF) test, where the null hypothesis (H0) assumes the presence of a unit root, indicating non-stationarity, while the alternative hypothesis (H1) suggests stationarity. For each time series type, we introduced gaps of varying sizes (5, 10, and 20) within the data and assessed the performance of our method using three evaluation metrics: sMAPE, RMSE, and MAE. The results are reported separately for the left and right hinges, which represent the regions before and after the introduced gaps, respectively. The findings highlight the effectiveness and robustness of Hinge-FM2I in handling gaps within univariate time series data, as evidenced by the reasonably low error values across different gap sizes and time series characteristics, regardless of their stationarity.

\begin{table}[H]
	\centering
	\begin{tblr}{
			row{2} = {c},
			row{4} = {c},
			row{5} = {c},
			row{7} = {c},
			row{8} = {c},
			cell{1}{1} = {r=2}{},
			cell{1}{2} = {r=2}{c},
			cell{1}{3} = {c=3}{c},
			cell{1}{6} = {c=3}{c},
			cell{3}{1} = {r=3}{},
			cell{3}{2} = {c},
			cell{3}{3} = {c},
			cell{3}{4} = {c},
			cell{3}{5} = {c},
			cell{3}{6} = {c},
			cell{3}{7} = {c},
			cell{3}{8} = {c},
			cell{6}{1} = {r=3}{},
			cell{6}{2} = {c},
			cell{6}{3} = {c},
			cell{6}{4} = {c},
			cell{6}{5} = {c},
			cell{6}{6} = {c},
			cell{6}{7} = {c},
			cell{6}{8} = {c},
			hline{1,3,6,9} = {-}{},
			hline{2} = {3-8}{},
			hline{4-5,7-8} = {2}{},
		}
		Time Series feature        & Gap Size & Left-Hinge &         &        & Right-Hinge &         &        \\
		&          & sMAPE      & RMSE    & MAE    & sMAPE       & RMSE    & MAE    \\
		Stationary Time Series     & 5        & 9.98       & 486.63  & 392.11 & 9.59        & 454.72  & 368.42 \\
		& 10       & 13.03      & 895.84  & 643.79 & 13.36       & 883.69  & 642.54 \\
		& 20       & 16.22      & 1024.57 & 773.52 & 15.92       & 1051.11 & 778.59 \\
		Non-Stationary Time Series & 5        & 5.95       & 277.04  & 226.63 & 5.72        & 270.50  & 219.10 \\
		& 10       & 8.33       & 400.26  & 322.74 & 8.75        & 434.45  & 346.52 \\
		& 20       & 10.31      & 520.02  & 415.18 & 10.71       & 552.09  & 439.87 
	\end{tblr}
	\caption{Comparison of the robustness of our approach on time series with various features}
	\label{Tab:Diff}
\end{table}

\begin{figure}[!ht]
	\centering
	\begin{subfigure}[b]{0.3\textwidth}
		\includegraphics[width=\textwidth]{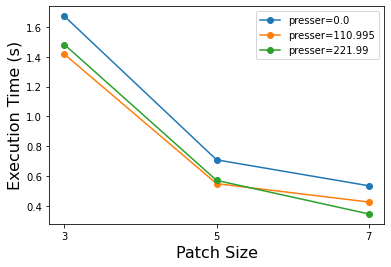}
		\caption{}
	\end{subfigure}
	\hfill
	\begin{subfigure}[b]{0.3\textwidth}
		\includegraphics[width=\textwidth]{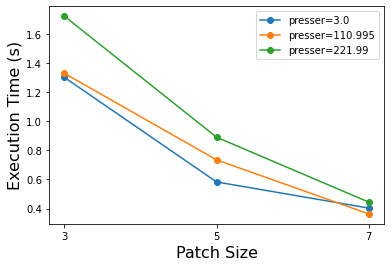}
		\caption{}
	\end{subfigure}
	\hfill
	\begin{subfigure}[b]{0.3\textwidth}
		\includegraphics[width=\textwidth]{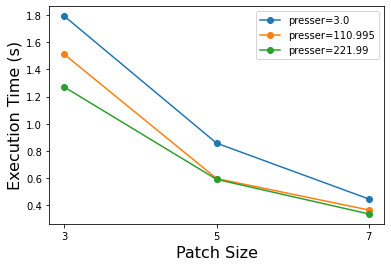}
		\caption{}
	\end{subfigure}
	
	\medskip
	
	\begin{subfigure}[b]{0.3\textwidth}
		\includegraphics[width=\textwidth]{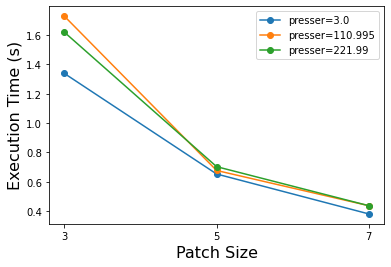}
		\caption{}
	\end{subfigure}
	\hfill
	\begin{subfigure}[b]{0.3\textwidth}
		\includegraphics[width=\textwidth]{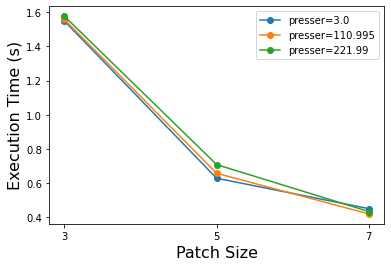}
		\caption{}
	\end{subfigure}
	\hfill
	\begin{subfigure}[b]{0.3\textwidth}
		\includegraphics[width=\textwidth]{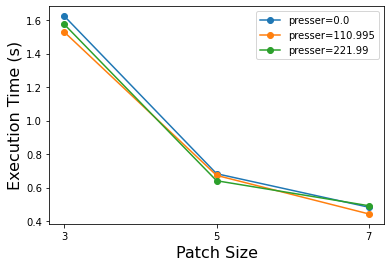}
		\caption{}
	\end{subfigure}
	
	\caption{Computational times for six different images transformation and their inpainting, based on patch size.}
	\label{fig:computational_results}
\end{figure}

Alongside evaluating the imputation accuracy of our proposed method, we further analyzed its computational efficiency across different configurations. As shown in Table \ref{Tab:time}, it reports the minimum, maximum, and mean execution times measured over 10 independent runs, considering variations in matrices, patch sizes, and presser. To provide deeper insights, we also showcase in Fig.\ref{fig:computational_results} different computational time of various transformation and their inpainting. We can see that using a higher patch size gives a decent computational time.

\begin{table}[H]
	\centering
	\begin{tblr}{
			cells = {c},
			cell{1}{1} = {r=2}{},
			cell{1}{2} = {c=3}{},
			cell{1}{5} = {c=3}{},
			cell{1}{8} = {c=3}{},
			hline{1,3-4} = {-}{},
			hline{2} = {2-10}{},
		}
		Method      & gap of size 5 &      &      & gap of size 10 &      &      & gap of size 20 &      &      \\
		& Min           & Mean & Max  & Min            & Mean & Max  & Min            & Mean & Max  \\
		Left\_hinge & 0.49          & 1.25 & 3.29 & 0.72           & 1.79 & 4.02 & 1.37           & 3.65 & 8.86 
	\end{tblr}
	\caption{Comparison of computational time of the proposed approach}
	\label{Tab:time}
\end{table}

\section{Conclusions and perspectives}
This study addressed the challenging problem of imputing large gaps of missing data in univariate time series. We proposed a novel method, Hinge-FM2I, which build upon the existing FM2I approach to select optimal replacements for large gaps of missing data. Extensive evaluations demonstrated the consistent superiority of our proposed methods over widely-used imputation techniques across various metrics, highlighting their effectiveness in choosing replacements. While the findings presented in this work offer valuable insights, there remain opportunities for further research exploration. One potential direction is the investigation of the Both Hinge-FM2I method, which leverages information from both sides of the missing gap, rather than relying solely on data from one end. Additionally, we could explore strategies for selecting diverse elements from multiple rows, instead of utilizing consecutive points from a single row. The substantial improvements in imputation accuracy achieved by Hinge-FM2I are highlighted in Table \ref{Tab:Best}. For instance, when imputing a gap of size 5, our method achieved a sMAPE of 6.4, significantly outperforming the worst possible choice sMAPE = 197.2 and even approaching the theoretically ideal but often impractical best choice sMAPE = 0.99. Furthermore, enhancing the computational efficiency of our approach through parallelization of matrix generation and inpainting processes presents an avenue for future work. Regarding the accuracy, a patch-based algorithm \cite{noufel2024hysim} has been introduced and shown promising results for image restoration and interpolation. It is under investigating for being fully integrated into Hinge-FM2I.

\begin{table}[H]
	\centering
	\caption{Comparison of Forecast Accuracy Metrics}
	\label{Tab:Best}
	\begin{tabular}{|l|l|l|l|}
		\hline
		\diagbox{Size}{Choice} & Worst Choice & Hinge method & Best choice \\
		\hline
		5 & 197.2 & 6.4 & 0.99 \\
		\hline
		10 & 196.33 & 10.57 & 0.62 \\
		\hline
		20 & 196.91 & 12.41 & 1.12\\
		\hline
	\end{tabular}
\end{table}

\printbibliography

\end{document}